\documentclass[sn-standardnature]{sn-jnl}%

\usepackage{xcolor}
\usepackage{float}
\usepackage{algorithm}
\usepackage{algpseudocode}
\usepackage{makecell}
\usepackage{graphicx}
\usepackage{tabularray}
\usepackage{lineno}
\usepackage[usestackEOL]{stackengine} %

\jyear{2023}%

\theoremstyle{thmstyleone}%

\theoremstyle{thmstyletwo}%

\theoremstyle{thmstylethree}%

\newcommand{\networkName}{UMedPT}
\newcommand{\networkNameFixed}{UMedPT-fixed}
\newcommand{\networkNameAffine}{UMedPT-A}
\newcommand{\embargonotice}{\textit{Due to an embargo this section will be updated after 01.01.24}}
\newif\ifpreprint
\preprinttrue

\begin{document}

\title[\networkName{} - Foundational Biomedical Pretraining]{\vspace{-2cm}Overcoming Data Scarcity in Biomedical Imaging with a Foundational Multi-Task Model}

\author[1]{\fnm{Raphael} \sur{Schäfer}}

\author[1]{\fnm{Till} \sur{Nicke}}%

\author[1]{\fnm{Henning} \sur{Höfener}}%

\author[1]{\fnm{Annkristin} \sur{Lange} }%

\author[1,3]{\fnm{Dorit} \sur{Merhof}}%

\author[4,5]{\fnm{Friedrich} \sur{Feuerhake}}%

\author[1,2]{\fnm{Volkmar} \sur{Schulz}}%

\author[1]{\fnm{Johannes} \sur{Lotz}}%

\equalcont{JL and FK contributed equally to this work.}

\author[1,2]{\fnm{Fabian} \sur{Kiessling}}%

\equalcont{JL and FK contributed equally to this work.}

\affil[1]{\orgname{Fraunhofer Institute} for \orgdiv{Digital Medicine MEVIS}}

\affil[2]{\orgdiv{Institute for Experimental Molecular Imaging}, \orgname{RWTH Aachen University}}

\affil[3]{\orgdiv{Institute of Image Analysis and Computer Vision, Faculty of Informatics and Data Science}, \orgname{University of Regensburg}}

\affil[4]{\orgdiv{Institute for Pathology}, \orgname{Hannover Medical School}}

\affil[5]{\orgdiv{Institute for Neuropathology}, \orgname{University Clinic Freiburg}}

\abstract{
Foundational models, pretrained on a large scale, have demonstrated substantial success across non-medical domains. However, training these models typically requires large, comprehensive datasets, which contrasts with the smaller and more heterogeneous datasets common in biomedical imaging. Here, we propose a multi-task learning strategy that decouples the number of training tasks from memory requirements. We trained a Universal bioMedical PreTrained model (UMedPT) on a multi-task database including tomographic, microscopic, and X-ray images, with various labelling strategies such as classification, segmentation, and object detection. The UMedPT foundational model outperformed ImageNet pretraining and the previous state-of-the-art models. For tasks related to the pretraining database, it maintained its performance with only 1\% of the original training data and without fine-tuning. For out-of-domain tasks it required not more than 50\% of the original training data. In an external independent validation imaging features extracted using UMedPT proved to be a new standard for cross-center transferability.
}


\maketitle

\section*{Main}
\label{sec:main}

Deep learning has started to revolutionize biomedical image analysis, owing to its ability to learn and extract useful image representations.
A widely-adopted approach for enabling deep learning in biomedical image analysis involves pretraining models on extensive natural image datasets, such as ImageNet-1K \cite{deng2009imagenet}, and subsequently either fine-tuning them or utilizing pretrained features directly for specific target tasks \cite{Kim2022TLSurvey,kather2019deep,mei2022radimagenet}. Fine-tuning leverages the pretrained model's weights as initial foundation, enabling accelerated training and enhanced performance even in situations with limited data. Alternatively, such foundational models can be kept frozen, with their features directly applied to biomedical downstream tasks. Despite requiring more computation time and data amounts, fine-tuning has firmly established itself as standard practice across a diverse range of downstream computer vision tasks, encompassing object detection and semantic segmentation, among others \cite{DBLP:journals/corr/abs-1902-07208, DBLP:journals/corr/WangPLLBS17}.

Driven by the recent trend of increasingly large pretrainings, the need for foundational models in biomedical imaging has become clear \cite{Moor2023, Willemink2022}.
However, effective pretraining of deep neural networks requires large amounts of annotated training data, which are often scarce in biomedical imaging \cite{Litjens_2017Survey}. While many public small and medium-sized datasets exist in the biomedical domain, there is no single pretraining dataset comparable to ImageNet.

Several methods have been proposed to address the data scarcity problem when pretraining deep networks for biomedical image analysis. One approach is to use self-supervised learning, which learns visual representations from unlabeled data by solving pretext tasks. However, clear performance gaps exist between self-supervised and label-supervised pretraining methods \cite{zhou2022generalized}.

Another approach is to use domain-specific supervised pretraining. For example, Zhou et al. \cite{zhou2022generalized} used a large text-labeled chest X-ray dataset to train universal representations for chest X-rays. They evaluated their approach on unseen datasets and found that their chest X-ray encoder outperforms ImageNet pretraining by up to 10\% accuracy when applied to other chest X-ray analysis tasks. Nonetheless, supervised pretraining can only be applied to domains where large amounts of training data are available, such as radiographs.

Mei et al. \cite{mei2022radimagenet} proposed to combine multiple medical classification datasets into one larger dataset and use it for pretraining deep networks for radiology tasks. They presented target tasks where this approach can achieve better results than ImageNet pretraining. However, this approach has limitations: it relies solely on classification labels, which may not capture all relevant information in medical images, and it requires the network to predict all classes in the combined dataset, even if they are unrelated or not meaningful for a given case.

Multi-task learning (MTL) promises to provide a solution to data scarcity in biomedical image analysis by enabling simultaneous training of a single model that generalizes well across multiple tasks \cite{MTLSurveyZhang}. It takes advantage of the many small- and medium-sized datasets in biomedical imaging, efficiently utilizing different label types and data sources to pretrain image representations which are applicable to all tasks, enabling deep learning for domains with sparse data. MTL has been applied to biomedical image analysis in various ways, such as training on multiple small- and medium-sized datasets from distinct tasks, specifically limited to classification \cite{mormont2020multi} or segmentation \cite{TransUNETMedChen}. Additionally, MTL has been used with multiple label types for individual images, demonstrating that sharing features across label types enhances task performance \cite{OneModelGraham2022}.

To combine multiple datasets with different label types for large-scale pretraining, we introduce a novel multi-task training strategy specifically designed to address the data scarcity problem in biomedical imaging by learning versatile representations across diverse biomedical modalities, diseases, and label types. To cope with the memory constraints commonly encountered in large-scale multi-task learning, our approach employed a gradient accumulation-based training loop whose scaling is almost unconstrained with respect to the number of training tasks. 

\begin{figure}[H]
    \centering
    \includegraphics[width=0.85\linewidth]{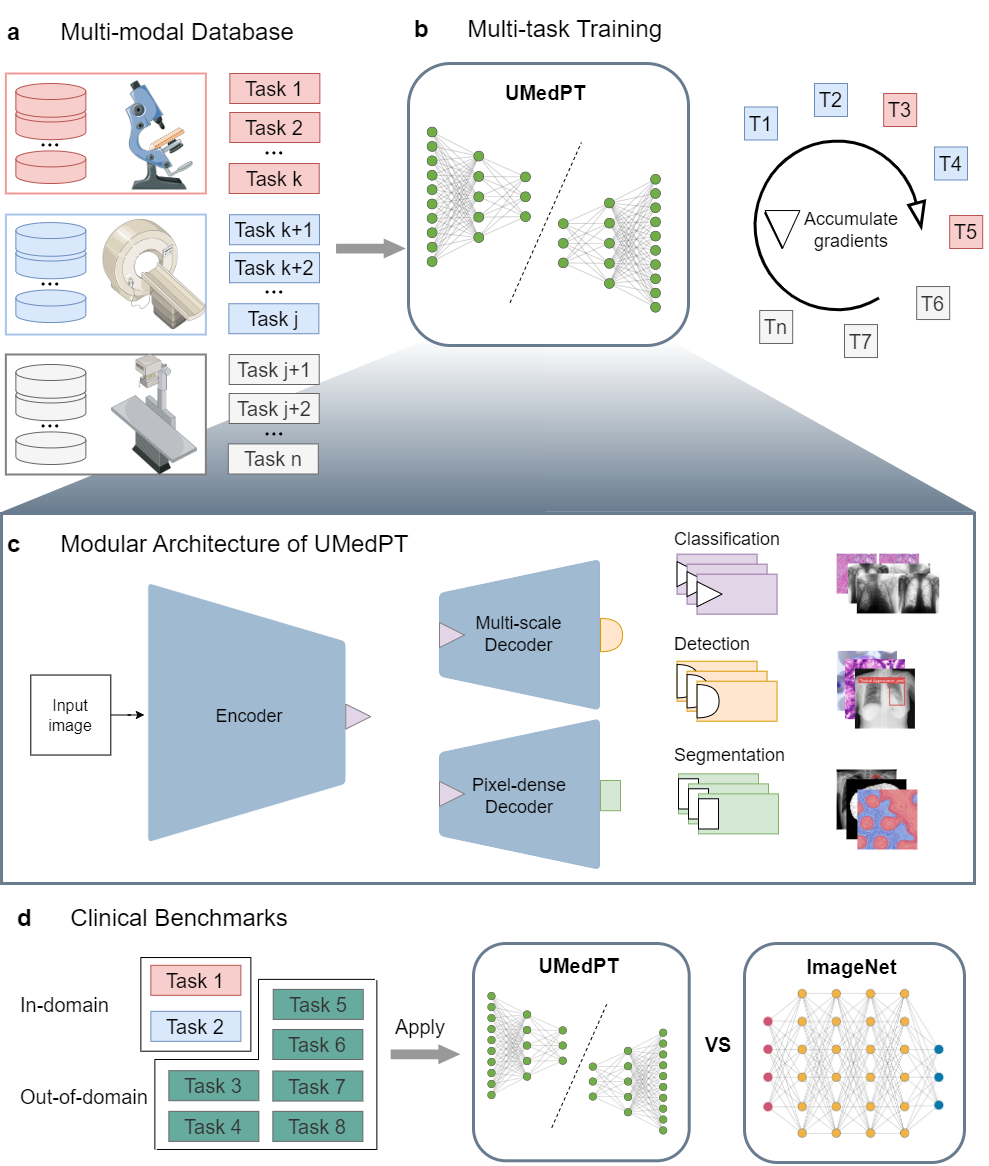}
    \caption{\textbf{Study Overview. }Illustration of the organization of the study and the multitask learning
approach:
\textbf{a} Heterogeneous data sources from histology (red), tomographic imaging (blue),
and X-ray (gray) formed the generalized training database.
\textbf{b} The pretraining process trained generally applicable neural network components (shared blocks). A novel training loop based on gradient accumulation, detailed in Algorithm \ref{alg:train}, allowed for the joint training incorporating all pretraining tasks at each optimization step despite different label types.
\textbf{c} Depending on the label type of the current task being processed, our modular architecture was assembled from a set of three shared blocks. More information about each shared block can be found in Figure \ref{fig:arch_overview}.
\textbf{d} The applicability of the pretrained neural components to medical imaging was evaluated using a diverse benchmark consisting of several clinically relevant small datasets categorized as in-domain or out-of-domain. The performance of \networkName{} was compared with that of ImageNet pretraining.
}
\label{fig:study_overview}
\end{figure}

Figure \ref{fig:study_overview} presents the architecture of our neural network which consisted of shared blocks, such as the encoder, a segmentation decoder, and a localization decoder, along with task-specific heads.
The shared blocks were trained for application to all pretraining tasks, facilitating the extraction of universal features, while the task-specific heads handled label-specific loss calculation and predictions. Our tasks included three supervised label types: object detection, segmentation, and classification. Classification tasks, for instance, can model binary biomarkers, segmentation tasks can extract spatial information, and object detection tasks can be used, for example, to train biomarkers based on cell quantities.

Using this architecture and training strategy, we developed, to our knowledge, the first fully supervised foundational model for biomedical imaging named \networkName{}, using 17 tasks based on 15 datasets and their original annotations.
Each task consisted of training and test sets with its label type, e.g., classification, segmentation, or object detection.
A study overview is presented in Figure \ref{fig:study_overview}.

\networkName{} consistently matched or outperformed the pretrained ImageNet network in in- and out-of-domain tasks, while maintaining strong performance with substantially less training data in both direct application of image representations (frozen) and fine-tuning settings.
Furthermore, we compared our model with external reference results and demonstrated the robustness of \networkName{} though external validation.

By providing the training framework and model as open-source software, we facilitate the training and implementation of data-efficient biomedical imaging models.
Serving as a basis for future advancements in data-scarce domains, \networkName{} opens perspectives to extend the application of deep learning in medical fields where collecting large cohorts is particularly challenging, such as rare diseases and pediatric imaging.

\section*{Results}
\label{sec:results}

We designed two benchmarks: the “in-domain benchmark” to assess the applicability of \networkName{} to problems closely related to its training database and the “out-of-domain benchmark” to evaluate its performance in unfamiliar domains.
Additionally, we compared our findings with previously published results from the same datasets being considered the state of the art.

Furthermore, we assessed \networkName{}'s performance in data-scarce scenarios by training it with varying amounts of original training data, ranging from 1\% to 100\%, and report the average results gathered from five repeated runs for each experimental setup.
All networks were trained with a frozen encoder and subsequently in a fine-tuning setting with the same training scheme and hyperparameters.

Ablation studies are included in Extended Data Tables \ref{tab:resultsweakcombined}, \ref{tab:resultsstrongfrozen} and \ref{tab:resultsstrongfinetune}. First \networkNameFixed{}, which consistently used an image size of (224, 224), and \networkNameAffine{}, which retains the same image dimensions but replaces static layernorms with trainable (affine) layernorms. In our evaluations across various tasks, \networkName{} outperformed ImageNet by an average of 8.5\% and surpassed \networkNameFixed{} by 2.97\%. Compared with \networkNameFixed{}, \networkNameAffine{} showed an average performance gain of 0.37\%.

\subsection*{For in-domain tasks, \networkName{} matches the baseline performance with 1\% of the data without fine-tuning}

\begin{figure}
    \centering
    \begin{minipage}[b]{1.0\textwidth}
        \stackinset{l}{1pt}{t}{1pt}{\color{black}\bfseries a}{\includegraphics[width=\textwidth]{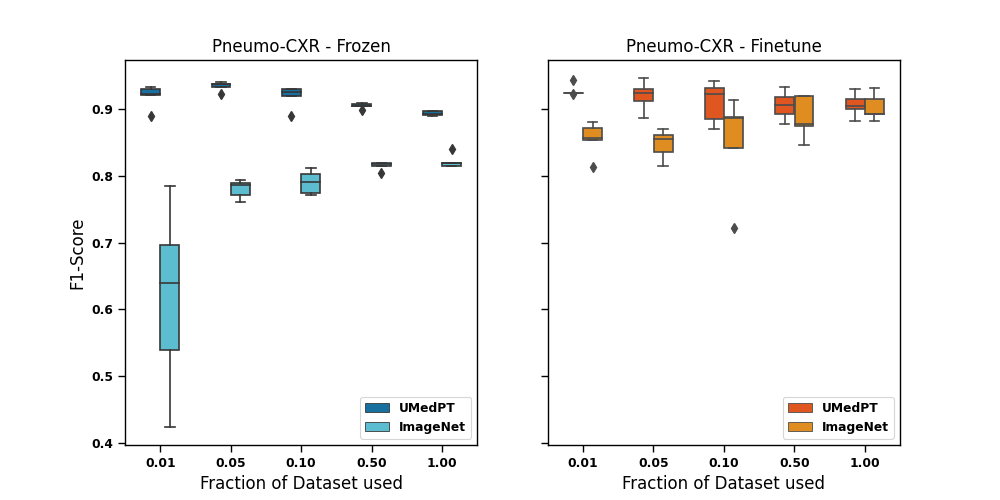}}
        \label{fig:resultsweak2}
    \end{minipage}
    
    \begin{minipage}[b]{1.0\textwidth}
        \stackinset{l}{1pt}{t}{1pt}{\color{black}\bfseries b}{\includegraphics[width=\textwidth]{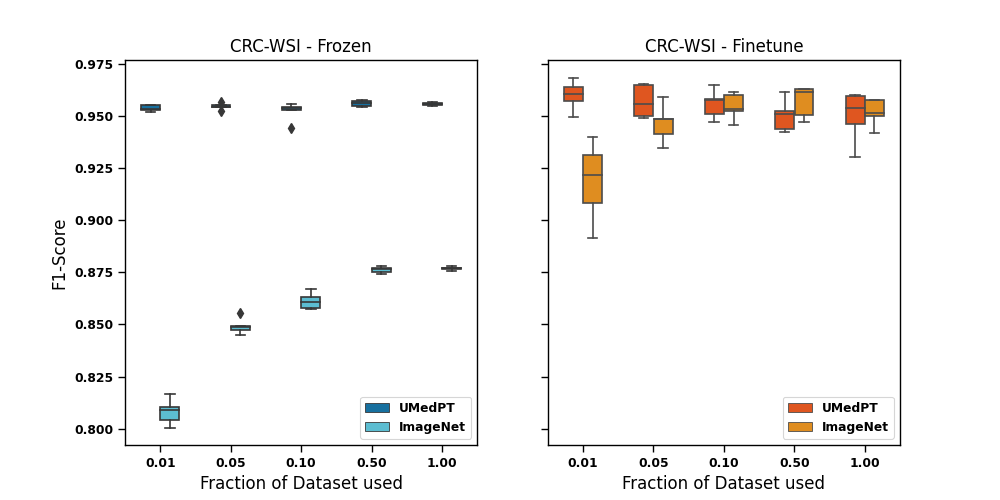}}
        \label{fig:resultsweak1}
    \end{minipage}
    
    \caption{\textbf{Results for in-domain tasks.} \textbf{a} In diagnosing tuberculosis (Tuber-CXR), \networkName{} matched the full fine-tuned performance of ImageNet, even with a frozen encoder and a reduced dataset size (50\%). \textbf{b} For BC-Bach-WSI, fine-tuning was essential for high performance. When the encoder was frozen, half the training data was required to match the result obtained with ImageNet in a similar setting.
    }
    \label{fig:resultsweak}
\end{figure}

We compared \networkName{} to results obtained with weights pretrained on ImageNet.
In both tasks, \networkName{} was able to match the best performance of the ImageNet baseline over all configurations using only 1\% of the original training data.
Notably, our model achieved higher performance with a frozen encoder compared to our model with fine-tuning, as shown in Figure \ref{fig:resultsweak}.
Below, we outline key findings in two in-domain tasks.

\paragraph{CRC tissue classification (CRC-WSI):}

In CRC-WSI, we classified colorectal cancer (CRC) tissue from microscopic whole slide images (WSI) into nine different classes, including adipose tissue, normal colon mucosa, and colorectal adenocarcinoma epithelium \cite{kather2019predicting}.

ImageNet achieved an average F1-score of 95.2\% in predicting the nine different tissue classes on the unseen test set using all of the training data and fine-tuning.
\networkName{} achieved a comparable performance with 1\% training data and a frozen encoder (95.4\% F1-score, see Figure \ref{fig:resultsweak}).

When the training dataset size was increased to 50\% and 100\% and the models were fine-tuned, the results converged to approximately the same F1-score across all methods (Extended Data Table \ref{tab:resultsweakcombined}).
Surprisingly, for \networkName{}, increasing the training data beyond 1\% did not enhance the model's performance and sometimes tended to degrade it. Notably, it did not matter which 1\% were picked as the final performance had a low variance.

\paragraph{Pneumonia in pediatric cohort (Pneumo-CXR):}

Pneumo-CXR focused on diagnosing pediatric pneumonia \cite{KERMANY20181122}. Here, \networkName{} outperformed ImageNet across all dataset sizes. The best performance of \networkName{} was achieved using 5\% of the data ($\approx$ 250 images) and frozen features, resulting in an F1-score of 93.5\%. The best ImageNet performance (90.3\% F1-score, 100\% of the data) was matched with the smallest split (1\% of the data, $\approx$ 50 images, see Extended Data Table \ref{tab:resultsweakcombined}).

\subsection*{For out-of-domain tasks, \networkName{} maintains ImageNet performance despite reduction in training data}

\begin{figure}
    \centering
    \begin{minipage}[b]{1.0\textwidth}
        \stackinset{l}{1pt}{t}{1pt}{\color{black}\bfseries a}{\includegraphics[width=\textwidth]{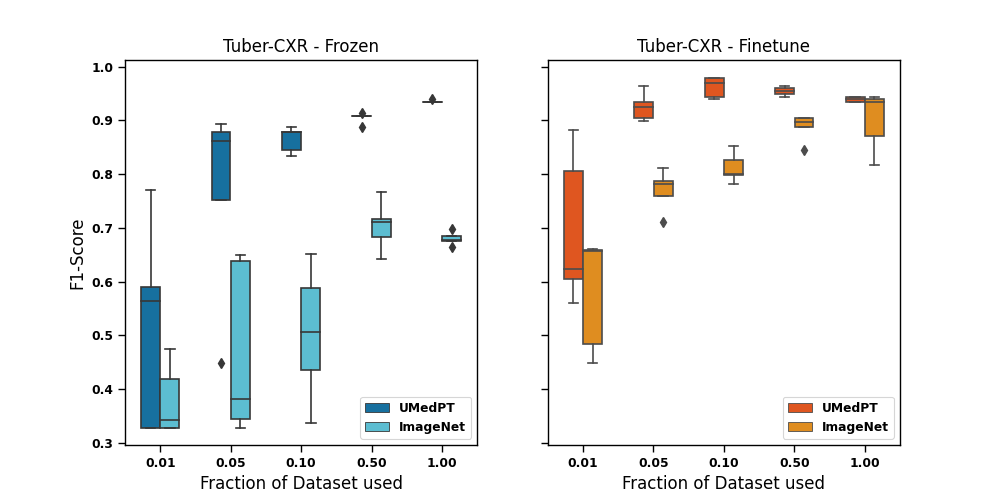}}
        \label{fig:resultsstrong1}
    \end{minipage}
    
    \begin{minipage}[b]{1.0\textwidth}
        \stackinset{l}{1pt}{t}{1pt}{\color{black}\bfseries b}{\includegraphics[width=\textwidth]{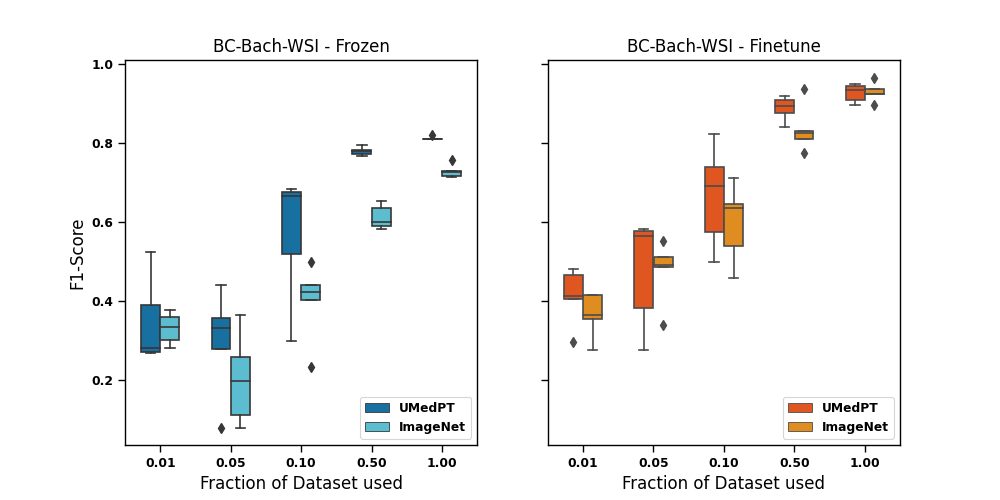}}
        \label{fig:resultsstrong2}
    \end{minipage}
    \caption{\textbf{Results for out-of-domain tasks.} \textbf{a} For tuberculosis diagnosis (Tuber-CXR), \networkName{} enables matching the full ImageNet performance even with frozen encoder and a reduced amount of data (50\%). \textbf{b} In BC-Bach-WSI, fine-tuning was necessary for a high performance. In the frozen setting, half of the training data was required to match the ImageNet frozen result.}
    \label{fig:resultsstrong}
\end{figure}

In the out-of-domain benchmark, \networkName{} compensated a data reduction of 50\% or more across all datasets in the benchmark when the encoder was frozen, as detailed in Extended Data Table \ref{tab:resultsstrongfrozen}.
With fine-tuning, ImageNet's performance consistently improved if more data were used. Here, for two out of five datasets, \networkName{} was able to match the performance of ImageNet using only 50\% or less of the data, even when fine-tuning was applied (see Figure \ref{fig:resultsstrong} and Extended Data Table \ref{tab:resultsstrongfinetune}). Below, we highlight key findings in the out-of-domain tasks:

\paragraph{Tuberculosis diagnosis in CXR (Tuber-CXR):}

In the task of diagnosing tuberculosis from chest X-ray (CXR) \cite{jaeger2014two}, \networkName{} delivered the highest average result. This was achieved by fine-tuning the model and using just 10\% of the data, resulting in an F1-score of 96.3\%. Adding more training data did not further improve the score for our model (Figure \ref{fig:resultsstrong}). To match the overall best average result of ImageNet, \networkName{} required 5\% of the data with fine-tuning and 50\% of the data with a frozen encoder.

\paragraph{CNS neoplasia diagnosis in MRI (CNS-MRI):}

We used the CNS-MRI \cite{BrainTumorMRIDataset2020} dataset to train our system to diagnose CNS neoplasms from MRI scans.
ImageNet, with frozen features, achieved an F1-score of 89.0\%. \networkName{} was able to match this score using 5\% of the training data.
With the full training set and fine-tuning, \networkName{} achieved the top F1-score of 99.3\%.

\paragraph{Breast cancer classification in WSI (BC-Bach-WSI):}

The BC-Bach-WSI dataset was used for breast cancer classification in whole slide images.
Using the frozen encoder, ImageNet achieved an F1-score of 72.9\%. \networkName{} obtained this score with 50\% of the data, resulting in an F1-score of 78.0\%.
Here, the best results were achieved using fine-tuning.

\paragraph{Breast cancer classification in microscopic images (BC-BHis-MIC):}

The BC-BHis-MIC dataset was used for breast tumor classification into benign and malignant in microscopic (MIC) images. The best mean ImageNet result was achieved with 100\% of the data and fine-tuning, resulting in an F1-score of 98.4\%. \networkName{} also achieved an F1-score of 98.4\%. When using a frozen encoder, ImageNet achieved an F1-score of 82.3\%. \networkName{} was able to match this score using 50\% of the data.

\paragraph{Organ segmentation in MRI (OrganSeg-MRI):}

The OrganSeg-MRI dataset \cite{ji2022amos} was used for organ segmentation in MRI scans. Training with the frozen encoder did not yield satisfactory results, neither for ImageNet nor for \networkName{}. The best average baseline result was observed when using 100\% of the data and fine-tuning, resulting in a mean Intersection over Union (mIoU) of 0.859. Even in this setting, \networkName{} outperformed ImageNet, achieving an mIoU of 0.881. \networkName{} needed 50\% of the data to match the best ImageNet result.

\subsection*{Comparison of \networkName{} with external reference results}

Next, we compared the performance of \networkName{} to outcomes reported in the literature.
When using the frozen encoder configuration, \networkName{} surpassed the external reference results in four out of six tasks. Notably, the two tasks where \networkName{} underperformed were out-of-domain (breast cancer classification BC-Bach-WSI and CNS neoplasia diagnosis CNS-MRI).

With fine-tuning, \networkName{} exceeded the external reference results across all tasks (see Table \ref{tab:datafraction}). 

A comparison for the OrganSeg-MRI task could not be performed, because no results specific to the MRI-only subtask of the challenge were reported. The reference results are further detailed in the supplement, Section \ref{sec:sotadetails}.

For the two in-domain target tasks, 1\% of the original training data were enough to reach the external reference results. For out-of-domain tasks \networkName{} required $\leq$ 50\% of the original training data to match the state-of-the-art performance when fine-tuning is used. With frozen features, in three out of four tasks \networkName{} required either 100\% of the data or did not match the state of the art.

\begin{table}[h]
    \caption{Amount of data required by \networkName{} to match state-of-the-art performance on classification tasks from different imaging domains. Datasets marked with an asterisk (*) were compared across different test splits.}
    \centering
    \begin{tabular}{llcc}
        \hline
        & & \multicolumn{2}{c}{\textbf{UMedPT}} \\
        \textbf{Task} & \textbf{Reference results}                        & \textbf{Frozen} & \textbf{Fine-Tuning}  \\ \hline
        CRC-WSI          & $>94\%$ accuracy \cite{kather2019predicting}   & $1\%$           & $1\%$                 \\ \hline
        Pneumo-CXR       & $92.8\%$ accuracy \cite{KERMANY20181122}       & $5\%$           & $1\%$                 \\ \hline
        Tuber-CXR*       & $82\%$ accuracy \cite{jaeger2013automatic}     & $10\%$          & $5\%$                 \\ \hline
        Tuber-CXR*       & $98.0\%$ AUC \cite{zhou2022generalized}        & $50\%$          & $10\%$                \\ \hline
        CNS-MRI*         & $>96\%$ accuracy                               & -               & $50\%$ \\ \hline
        BC-Bach-WSI*     & $87\%$ accuracy \cite{ARESTA2019122}           & -               & $50\%$                \\ \hline
        BC-BHis-MIC* & $88\%$ F1-score \cite{chhipa2022magnification} & $100\%$         & $10\%$                \\ \hline
    \end{tabular}
    \label{tab:datafraction}
\end{table}

\subsection*{\networkName{} provides robust image representations across clinical centers}

\ifpreprint
\embargonotice{}
\fi

\section*{Discussion}

Using a novel multi-task training strategy, \networkName{} fuses knowledge from multiple sources to compute robust image representations across diverse biomedical modalities, diseases, and label types.
We demonstrate that this knowledge can be effectively transferred to unseen target tasks.
Similarly, previous studies reported that pretraining on a large dataset \cite{zhou2022generalized} or multi-task pretraining \cite{mormont2020multi} can improve models' performance on small, unseen datasets.
In the field of medicine, this is, for example, especially important for rare and pediatric diseases, where the quantity of available images is often too small to effectively train deep neural networks.
However, the performance advantage of \networkName{} in in-domain tasks compared to out-of-domain tasks indicates that it is not entirely universal for all biomedical imaging applications yet, necessitating a broader scale of training.
The extent to which such pretraining should be scaled remains an open question. In natural language processing, larger models have demonstrated improvements with scaling, with some studies using up to 540 billion parameters \cite{singhal2023large}.
However, even non-biomedical foundational vision models, which are not limited by data scarcity, are 10 to 1000 times smaller than their natural language counterparts \cite{radford2021learning, kirillov2023segment}. This suggests that large vision models may not require an excessive number of parameters. Nevertheless, larger models typically require larger datasets, and our approach of combining labeled datasets into a single training could facilitate scaling of foundational vision models.

Alternatively, self-supervised pretraining can be applied. However, recent literature has proposed that label-supervised pretraining for imaging typically outperforms self-supervised pretraining empirically \cite{zhou2022generalized, oliver2018realistic} and theoretically \cite{castro2020causality}. Nonetheless, it still offers value in regularizing models and might help in reducing the volume of labelled data necessary.
Our approach can be extended to include an arbitrary number of self-supervised tasks into the pretraining, which may further enhance the generalizability of \networkName{}, especially in domains where abundant data are available but labelling is difficult or costly.

Training AI models from scratch can be computationally intensive. Here, foundational models such as \networkName{} in a frozen configuration may enable efficient feature extraction without additional training. Frozen features from pretrained networks solved in-domain classification tasks in pathology  \cite{mormont2020multi} and radiology \cite{LungGANs2023, arun2023pediatric}.
For natural images, the authors of CLIP \cite{radford2021learning} mentioned strong performance when directly reusing their frozen model for unseen tasks \cite{radford2021learning}.
However, the performance declined when frozen models were used for out-of-domain evaluations.
Our in-domain benchmark results are consistent with these findings, suggesting that frozen features should be the primary consideration for in-domain tasks. Moreover, 
we demonstrate that a single network can effectively extract features across multiple domains,
extending the applicability of frozen features within the in-domain context. 

We show that for out-of-domain tasks, fine-tuning can outperform the frozen configuration if there are enough data or significant differences between the target task data distribution and the pretraining distributions.
Other multi-task studies \cite{mormont2020multi} have observed that fine-tuning multi-task models or pretrained ImageNet models yield comparable performance.
However, even with fine-tuning, ImageNet did not outperform \networkName{} in any medical applications evaluated. In fact, \networkName{} showed advantages in the full data scenario with fine-tuning in two out-of-domain tasks. This could potentially be due to either the larger scale of \networkName{}'s training, which resulted in a well-generalizing base for fine-tuning, or the possibility that these two out-of-domain datasets were not sufficiently large at full size, making them ideal use cases for \networkName{}.

Differences in imaging modalities, scanners, annotations, or patient populations can make models fail when applied to data of different clinical centers \cite{10.1093/jamia/ocaa341}. Foundational models should be robust to multi-center variances, thereby improving their ability to generalize.
\ifpreprint
\embargonotice{}
\fi

Medical images vary in size, challenging deep learning methods that typically require uniform sizes. Homogeneous downsampling, however, resulted in reduced performance when comparing \networkName{} and \networkNameFixed{}, which is consistent with previous findings \cite{TransUNETMedChen}. For tasks that benefit from a large image size, our training method allows flexibility in choosing the optimal image size for each task, thus avoiding the problem of pre-defining it.

In conclusion, we demonstrated that the knowledge transfer capabilities of \networkName{} reduce the amount of data and time needed to train models for unseen tasks. In addition, \networkName{} can effectively manage domain shifts, thereby improving diagnostic performance in unfamiliar domains, such as pediatric cohorts or cohorts of unknown ethnicity.
Our results show that direct application of \networkName{} in a frozen setting can save time and computational resources while maintaining robustness across research centers and achieving state-of-the-art performance.
Even with large data and fine-tuning, we found no cases where ImageNet performed better than \networkName{}.
Thus, we hypothesize that \networkName{} is a valuable foundational model for biomedical imaging, especially when data are scarce.
Both the pretrained model and the multi-task training framework are available on Github (\url{https://... url-will-be-added-in-final-version}).

\section*{Methods}
\label{sec:methods}

In deep learning, training of models is performed by optimizing an objective function (loss function) that measures the difference between ground-truth labels and the result of the current models' iteration. The gradient of the loss function determines the extent of adjustments needed for each model parameter.

In the presented multi-task learning framework, the overall loss of the model was defined as the sum of the losses of all tasks that were evaluated simultaneously.

Every label type required the definition of a task-specific architecture component and an objective function computing its loss.
For \networkName{}, we combined these different components into a single model to solve many pretraining tasks simultaneously. These pretraining tasks integrated a large variety of the publicly available biomedical image data, including their annotations, into a single foundational training. This training lead to a shared model compatible to all of the pretraining tasks.

We addressed challenges such as memory constraints, varying input sizes, and label types to integrate a diverse set of small and medium-sized datasets for training \networkName. The model's design accommodated a variety of task types, including classification and dense-prediction tasks like segmentation, and allowed each task to operate using its optimal patch size and resolution, respectively.

\subsection*{Multi-task training strategy}\label{subsec:trainingstrategy}

A limiting factor in scaling currently established multi-task learning approaches is the increasing memory demand that is proportional to the number of tasks. This memory need is caused by processing all tasks in parallel during a single network pass.
To address this challenge, we developed a novel training strategy for \networkName{} that mostly decouples the number of training tasks from the memory requirements.

Our strategy achieved this by establishing an independent architecture, or 'computational graph', for each task. This graph is dynamically constructed and stored only during the active computation stage of each task.
To combine the individual graphs, we implement gradient accumulation (GA) before the optimization step as described in Algorithm \ref{alg:train}. GA allowed us to establish a training scheme wherein a single update step can consist of heterogeneous tasks in any order.

We ensure that the model's weights and gradients are stored only once, rather than duplicating them for each task. Additionally, only the activations for one task are kept in memory at a time. They are discarded after each task's backward pass, rather than stored for all tasks simultaneously. Therefore, the only memory requirements that increase with the number of tasks are related to the gradients of each task's head. As the shared section of the model represents the majority of the total model size, this approach allows for multi-task learning across many tasks, even on hardware with limited computational power.

Unlike traditional training schemes which merge multiple tasks in a batch of samples, GA enables flexible task scheduling. Each optimization step can therefore consist of multiple tasks and even multiple instances of the same task, enhancing the versatility of the proposed training approach.

When training a model on many tasks, the size of the respective dataset should not implicitly influence the task's weight in the overall model. To accomodate datasets of different sizes, we implemented an "infinite task sampler", which yielded one training batch of each task for every optimization step.
Once all data points of a task were used, the task sampler independently restarted the data loading for this task.
No information on the dataset's length was needed beforehand, which allowed each epoch to have a different length depending on data augmentation.
When training a multi-task model with GA, the model parameters were updated according to the sum of the model's gradients. Because summation is commutative, the order of tasks within an optimization step does not affect the outcome.

Classification, segmentation, and detection tasks have different loss functions with greatly varying magnitudes. This can lead to a situation where the loss of one task dominates the combined loss, resulting in insufficient training of other tasks.
To counteract this, we normalized the respective loss functions for each task type such that the expected value for random inputs for re-initialized weights was close to one.

We uniformly used the AdamW optimizer \cite{loshchilov2019decoupled} for all parameters. Following the training settings of \cite{liu2021swin} for ImageNet-1K training, we used a learning rate of 0.001 and a weight decay of 0.05 for all transformer-based models.

\begin{algorithm}
    \caption{The training loop processed tasks and their associated batches in the order given by the task sampler.
        For each step, a gradient was computed by evaluating the objective function of one task with respect to one of its batches.
        These gradients were accumulated until the task sampler initiated an optimization step.
        At this point, model parameters were updated considering all tasks since the last update, utilizing the accumulated gradient. After this, the gradient buffer was cleared for the next cycle.}
    \label{alg:train}
    \begin{algorithmic}[1] %

        \Procedure{Train}{$stepsPerEpoch$, $sharedblocks$, $tasksampler$, $optim$}
        \State $prepareSharedBlocks()$ \Comment norm \& task-specific modules

        \State $optim.clearGradients()$

        \For{$step \gets 0$ to $stepsPerEpoch$}
        \State $(batch, task) \gets tasksampler.next()$

        \State $loss \gets task.computeLoss(batch, sharedblocks)$
        \State $loss.backward()$ \Comment Accumulate gradients

        \If {$isUpdateStep(step)$} \Comment{E.g. after processing each task}
        \State $optim.updateParams()$
        \State $optim.clearGradients()$
        \EndIf
        \EndFor
        \EndProcedure
    \end{algorithmic}
\end{algorithm}

\subsection*{Architecture}\label{subsec:architecture}

Our tasks vary in their label types, each requiring problem-specific architectures. Therefore, we used a fixed-size embedding for classification tasks, designed to encapsulate features that are useful across all tasks. For segmentation tasks, we implemented a U-Net-like encoder-decoder scheme to learn multi-scale features. Additionally, object detection using FCOS \cite{tian2019fcos} required features produced by a feature pyramid network \cite{lin2017feature}.

To avoid wasting resources as not all features are required for every task, we created a modular architecture. We hypothesized that parameters should be largely shared across tasks. To address this, we placed most parameters into a shared encoder. To compute the necessary types of features, we then developed a pixel-dense decoder and a multi-scale decoder.

Our architecture supported encoders with configurable settings for image embedding dimensionality, stride (modulating the spatial range of feature pyramid levels), and feature pyramid depths. Given these settings are common in computer vision, our framework was able to integrate open-source encoder architectures. For segmentation tasks, we used a pixel-dense decoder that upsampled the feature pyramids in a U-Net style to generate pixel-features. For object detection, a multi-scale decoder was used to create feature maps from every pyramid level.

For the classification tasks, we directly used the image embedding from the encoder. This was computed using global average pooling from the lowest level of the feature pyramid, allowing us to handle variable input sizes. This approach is consistent with the ImageNet baseline. Segmentation tasks employed the shared pixel-dense decoder, while object detection tasks processed the encoder's output via a shared feature pyramid network. The output for each label type was computed using a single linear layer, or a single convolutional layer for dense prediction tasks.

In general, our proposed training loop can be used with any neural network, and \networkName{}'s decoders are compatible with any encoder that can generate multi-scale feature maps. We chose Swin Transformers as the encoder for \networkName{} \cite{liu2021swin}. They introduced a shifted windowing scheme that improved the efficiency of vision transformers with respect to image input size.

Regardless of the chosen architecture, normalization has been shown to be essential for accelerating the training process \cite{DBLP:journals/corr/IoffeS15, ba2016layer}. Batch normalization  \cite{DBLP:journals/corr/IoffeS15} is a widely used normalization technique. However, in our experiments, batch normalization led to poor performance (results not shown). One assumption when using batch normalization is that all input batches follow a common distribution which can be parameterized by a common statistical distribution. When combining different tasks and datasets, this assumption no longer holds.

While we observed that normalization layers enhanced training speed, they negatively impacted generalization within our training strategy. We found that the issue only arose when normalization statistics were computed across different tasks. To address this problem, we recursively replaced the original normalization layers in all shared blocks with layer normalization \cite{ba2016layer}, which by design do not require inter-task computation. This modification enabled the model to concurrently generalize across multiple training tasks.

We empirically analysed the effect of using layernorms with affine parameters in an adaptation \networkNameAffine{}.
\networkNameAffine{} was trained under the same setting as \networkNameFixed{}.

\begin{figure}
    \centering
    \includegraphics[width=\linewidth]{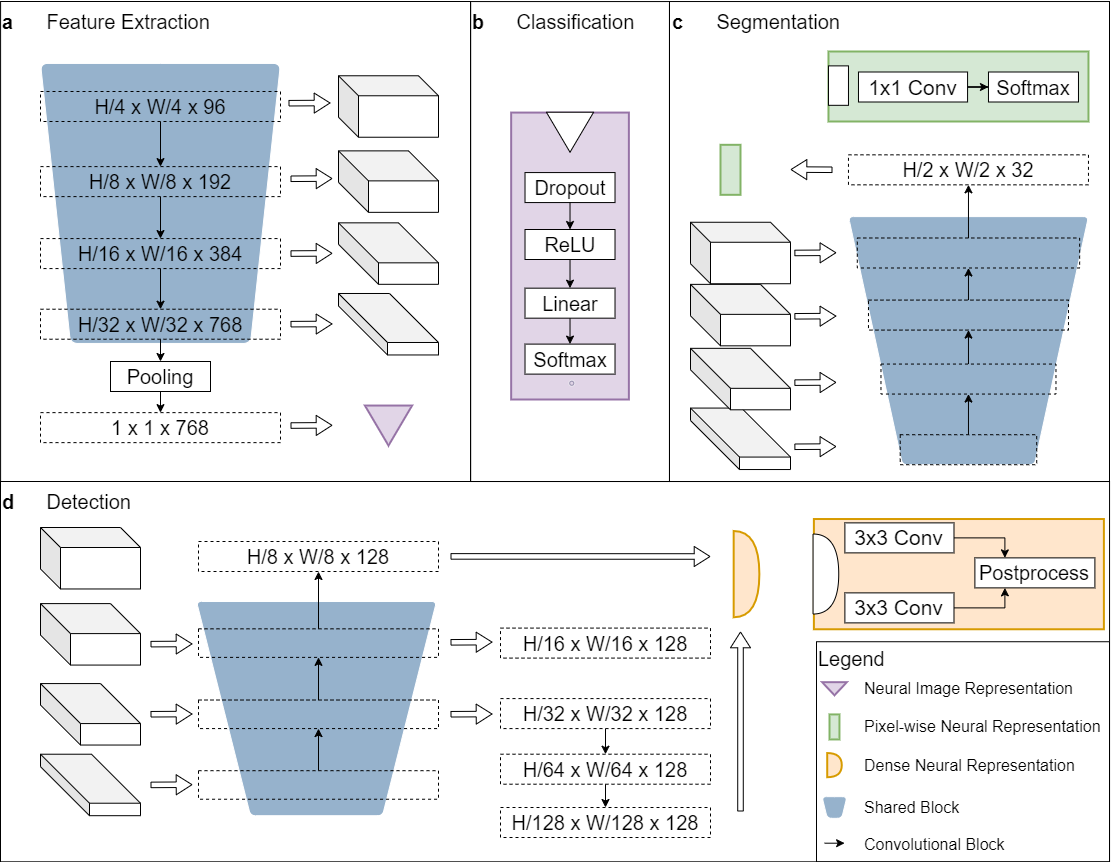}
    \caption{\textbf{\networkName's architecture.} \textbf{a} Features are extracted from an image of size HxW through a shared encoder. \textbf{b} Classification heads take the neural image representation of an image and apply a single linear layer. \textbf{c} A shared pixel-wise decoder processes the multi-scale feature maps and returns an embedding per pixel. Segmentation heads employing the popular U-Net spatial decoding strategy for handling segmentation features generate the prediction. \textbf{d} The multi-scale decoder uses the feature maps and transforms them into features for box-regression and box-classification. The FCOS-based detection head generates the final prediction.}
    \label{fig:arch_overview}
\end{figure}

\subsection*{Data loading from diverse sources}\label{subsec:dataloading}

\begin{figure}
    \centering
    \includegraphics[width=\linewidth]{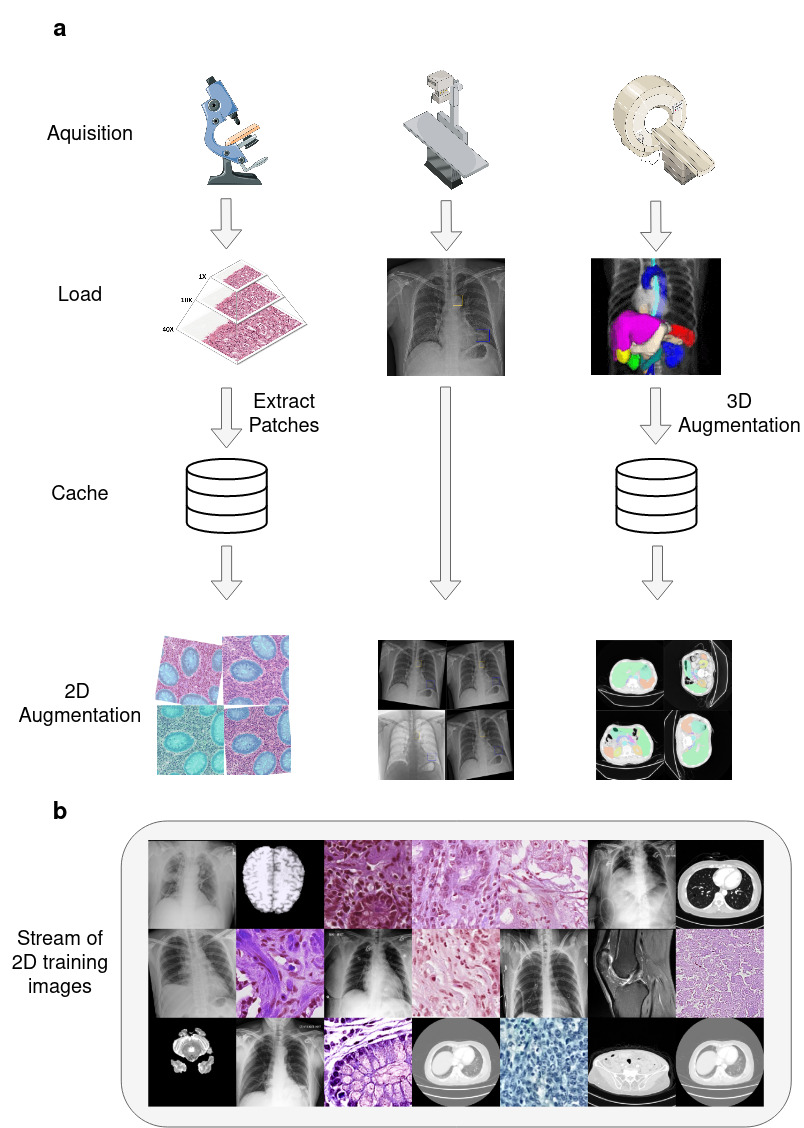}
    \caption{\textbf{Data pre-processing.} \textbf{a} We have categorized all of our tasks into one of three common formats: 2D images, 3D tomographic images, and gigapixel images. For each format, we developed a data loading strategy that transforms the data into the required 2D format. Additionally, for each domain, we implemented a standard augmentation strategy to be applied to the corresponding 2D images.
    \textbf{b} Our preprocessing results in a diverse stream of data including samples from all tasks. 
    }
    \label{fig:dataloading}
\end{figure}

To evaluate the ability of the model to learn multi-modal representations, we integrated a diverse range of biomedical imaging data types --- including microscopic pyramid gigapixel 2D images, standard 2D images (both grayscale and color), and 3D tomographic images --- into a single network. Each data type requires unique pre-processing and domain-specific augmentations for a universally applicable deep learning solution.
To accommodate these different data types, the encoder of \networkName{} used a standardized 2D image input format. This required the conversion of each data type into a two-dimensional format, as illustrated in Figure \ref{fig:dataloading}.

To ensure compatibility between different data types, we normalized all pixel intensities to a range between $0$ and $1$. For images with values ranging from 0 to 255, we divide them by 255. For 3D images, we normalize the maximum in the volume to 1 and the minimum to 0 for CT, and for MRI the intensity quantiles (2.5\%, 97.5\%) were used.

Smaller two-dimensional images from modalities like X-ray imaging, were resized to an edgelength of around 512 pixels. For larger images from histopathology, we extracted patches of similar size. Three-dimensional volumes were cut into slices to adapt them to the 2D format with task-specific patch sizes ranging from 224 to 512. Images did not need to be square for our training strategy.

We used a caching component to load whole slide images and 3D volumes. This method eliminated the need for pre-extraction of images, thereby enhancing data diversity. 3D augmentations were applied during every initial loading, while patch augmentations occured when a patch was retrieved from the cache.

\paragraph{Pretraining Tasks}

We selected 15 publicly available datasets for pretraining and extracted 17 tasks from them. We prioritized datasets that demonstrated acceptable performance in a single task setting. We defined acceptable performance as either aligning with the metrics reported by the dataset creators, when feasible, or passing a plausibility check conducted by a medical expert. Following these selection criteria, we included the following datasets in our study:

\begin{itemize}
    \item Amos22-CT organ segmentation in CT \cite{ji2022amos}
    \item Conic-WSI cell detection \cite{graham2021lizard}
    \item PICAL-MRI prostate cancer classification \cite{PICAI_BIAS}
    \item Panda-WSI prostate tissue semantic segmentation \cite{bulten2022artificial}
    \item Panda-WSI prostate tissue classification \cite{bulten2022artificial}
    \item VinBigData-CXR Thorax pathology pathology detection \cite{vinbigdata-chest-xray-abnormalities-detection}
    \item Crag-WSI Colorectal tissue semantic segmentation \cite{DBLP:journals/corr/abs-1806-01963}
    \item Brats2020-MRI brain semantic segmentation \cite{menze2014multimodal}
    \item CRC-WSI colorectal multi-class classification \cite{kather2019predicting}
    \item Avaniti-WSI prostate multi-label classification \cite{arvaniti2018automated}
    \item Cyto-WSI bone marrow single cell multi-class classification \cite{matek2021highly}
    \item Chexpert-CXR Thorax pathology multi-label classification \cite{irvin2019chexpert}
    \item SIIM-CXR pneumothorax semantic segmentation \cite{langershih2020}
\end{itemize}

In addition, we included the following auxiliary datasets for the purpose of meta-learning.
Unlike the other pretraining tasks, they did not focus on resolving a specific clinical issue:

\begin{itemize}
    \item ImageNet real world image classification \cite{deng2009imagenet}
    \item RadImageNet radiology multi-class classification \cite{mei2022radimagenet}
    \item COCO real world semantic segmentation \cite{lin2015microsoft}
    \item COCO real world object detection \cite{lin2015microsoft}
\end{itemize}

\subsection*{Data augmentation}

For 3D tomographic images, we applied standard 3D augmentations, followed by slicing and a set of standard domain specific 2D augmentations.
We augmented the orientation of the volume if the maximum edge length was more than two times the shortest edge length as proposed in \cite{isensee2018nnunet}.

For 2D images, we used standard augmentations using the albumentations library \cite{2018arXiv180906839B} (CLAHE, Sharpen, Emboss, RandomBrightnessContrast, RandomGamma, Gaussnoise and HueSaturationValue, ShiftScaleRotate).
For X-ray images we added image inversion with a 30\% probability, for histological imaging flips and channel shuffle and for MRI images flips and mirroring.

\subsection*{Task types}

\subsubsection*{Classification}

In the classification task, we handled data where a single input was associated with a single classification label from a set of $C$ classes. We used the latent representation computed by the encoder and processed it through a fully connected layer to obtain classification scores.

For classification tasks, we employed categorical cross-entropy $\mathcal{L}_{CCE}$ to compute the loss as implemented in \cite{paszke2019pytorch}. Typically, the loss magnitude would increase with a larger number of classes $C$. To prevent a bias towards tasks with more classes, we added a normalization term $\log C$ to compute the final loss as

\begin{equation}
    \mathcal{L}_{\text{multiclass}}(\hat y, y) = \mathcal{L}_{CCE}(\hat y, y) \cdot \log C,
\end{equation}

where $\hat y \in \mathbb R^C$ and $y\in \mathbb \{0, 1, ..., C\}$ are the true and predicted labels, respectively.

We did not utilize label smoothing or class probabilities in the classification task.

For multi-label classification tasks, we considered inputs that each had multiple binary classification targets $y$. In these cases, we chose the binary cross-entropy loss $\mathcal{L}_{BCE}$ from \cite{paszke2019pytorch}. To normalize the loss to $1$, we multiplied with a constant factor $\log_2(\exp(1)) \approx 1.44269$:

\begin{equation}
    \mathcal{L}_{\text{multilabel}}(\hat y, y) = \mathcal{L}_{BCE} \cdot 1.44269
\end{equation}

\subsubsection*{Semantic segmentation task}

The U-net architecture consisted of an encoder and a decoder, with the decoder producing dense pixel-level embeddings as output. To generate the final output, skip connections were established between the encoder's feature maps and the decoder's upsampling layers. These skip connections were implemented by concatenating the corresponding feature maps with the decoder's upsampling outputs. For training, we adopted the U-net decoder from \cite{Iakubovskii:2019}, which is an implementation of the original U-Net formulation proposed by \cite{ronneberger2015unet}.

The semantic segmentation task aimed to assign a class label to each pixel within an image, with the targets consisting of classes ranging from $1$ to $C$. For \networkName{} we configured its encoder to yield feature maps with strides of 4, 8, 16 and 32.

We employed an equally weighted combination of Dice loss and Focal loss for all segmentation tasks in \networkName. This strategy was chosen because Dice loss has been shown to perform well for mildly skewed datasets \cite{Jadon_2020}, while Focal loss is particularly effective for highly imbalanced datasets \cite{lin2018focal}. Thus, this combination allowed us to address challenges associated with both balanced and imbalanced datasets, such as the presence of large background regions, without the need for hyperparameter tuning.

\subsubsection*{Object detection task}

FCOS (Fully Convolutional One-Stage) \cite{tian2019fcos} is an anchor-free object detection method, which makes it an ideal candidate for our multi-task learning approach due to its similarity with our segmentation and classification methods, enabling efficient feature reuse.

Instead of traditional anchor-box based detection, FCOS employed pixel-dense predictions and a box-postprocessing technique. The architecture incorporated a shared encoder and a detection-specific decoder that produced two branches: one for classification and another for bounding box regression. The classification branch managed multi-class classification and centerness per pixel, while the bounding box regression branch predicted rectangle parameters, specifically the distances from each pixel to the edges of the bounding box. Although centerness is not essential to the algorithm, it helps suppressing low-quality bounding boxes. The final score for a box was computed by multiplying the predicted center-ness with the corresponding classification score.
To ensure homogeneity in the magnitudes of all multi-task losses and facilitate multi-task learning, we normalized the classification loss by dividing it by the number of classes $C$. This resulted in the following combined loss function: $\mathcal{L}_\text{objectdetection} = \mathcal{L}_\text{classification} \cdot \frac{1}{C} + \mathcal{L}_{\text{regression}} + \mathcal{L}_\text{centerness}$.

To reconstruct bounding boxes, the network produced a classification score per pixel, a centerness value, and rectangle parameters. Centerness was shared among tasks, fostering efficient multi-task learning, while each detection task employed one convolutional layer for classification  and one convolutional layer for regression. Similar to our segmentation task, the forward pass of the detection task generated one feature map for each downsampling step, typically resulting in five feature maps. These multi-level feature maps encapsulated spatial and semantic information at multiple resolutions, enhancing the method's efficacy in object detection and enabled encoding difficult ground truth cases involving overlapping and variably sized boxes.

\subsection*{Clinical validation}

\networkName{} was clinically validated using a diverse set of clinically relevant tasks. Our evaluation centered on two main aspects: the model's skill generalizability to new tasks and its proficiency in retrieving previously learned knowledge. These aspects were evaluated using two distinct benchmarks. We developed the downstream training schedule and tuned the hyperparameters using a simple synthetic dataset and ran the clinical evaluation exactly once without further hyperparameter tuning. For this reason, we did not use a validation set in our experiments.

The "in-domain" benchmark tested the model's ability to recall and adapt learned skills to new tasks. The "out-of-domain" benchmark measured the model's ability to adapt its learned skills to tasks distinct from the ones in the pretraining database.

Two distinct usage settings were considered in our evaluation: frozen and fine-tuning. In the frozen scenario, we directly extracted image representations from the shared blocks, thereby showing the usefulness of the learned representations.
Subsequently, the fine-tuning stage enabled the training of the shared blocks. The learning rates for the shared blocks in the transformer were set at $10^{-5}$, while the task-specific sections of the target tasks were trained at learning rates of $10^{-4}$.

To simulate data-scarcity and evaluate sample efficiency, we took multiple samples from the original training set at sizes of 1\%, 5\%, 10\%, 25\%, 50\%, and 100\%.
We utilized five splits of the training data to account for random selection effects, and ensured that all data from smaller splits also appear in the corresponding larger splits. Each method was evaluated for exactly the same five random train-test-splits. In consequence, a paired t-test was applied (significance level $p<0.05$), treating each baseline-\networkName{} result on a single train/test split as a pair.

\subsubsection*{In-domain benchmark}\label{sec:weakbenchmark}

We formulated a benchmark that aimed at examining the recoverability of knowledge encapsulated in \networkName. We selected knowledge already present in the pretraining database and examined their re-discovery potential by measuring the number of samples needed for re-identification, and compared the outcome with the ImageNet baseline on novel images.

\paragraph{CRC tissue classification (CRC-WSI):}

The CRC-WSI dataset consisted of Hematoxylin and Eosin (HE) stained images with nine tissue classifications that are largely balanced. The training set comprised 100,000 images extracted from 86 whole slide images, while the test data came from 25 different CRC patients. The label type of the CRC-WSI task was multi-class classification.

\paragraph{Pneumonia in pediatric cohort (Pneumo-CXR):}

The Pneumo-CXR dataset \cite{KERMANY20181122} consisted of pediatric chest X-rays, each labelled as either normal or pneumonia. Consistent with our approach for all datasets, we preserved the original label imbalance when downsizing the training datasets. The training set contained 1,349 normal cases and 3,883 pneumonia cases, while the test set contained 234 normal cases and 390 pneumonia cases. We treated Pneumo-CXR as a multi-class classification task with two classes.

\subsubsection*{Out-of-domain benchmark}\label{sec:strongbenchmark}

Here, we evaluated the transfer effectiveness of our method across a variety of clinically relevant tasks by establishing an out-of-domain generalization benchmark. In this benchmark each task's domain had to be different from all domains of the pretraining tasks.
We were able to increase the certainty of this claim by including only supervised pretraining tasks. Since the problem setting for each sample in a supervised task is known, this approach reduced the likelihood of pretraining knowledge overlapping with the benchmark.

\paragraph{Tuberculosis diagnosis in CXR (Tuber-CXR):}

The Tuber-CxR dataset \cite{jaeger2014two} consisted of postero-anterior (PA) chest radiographs that we used for multi-class classification. We randomly divided the images into a training set (70\% of the data) and a test set (30\% of the data) prior to any evaluation. The training set contained 239 tuberculosis samples and 225 normal samples, while the test set contained 51 tuberculosis samples and 101 normal samples.
This dataset was used as a multi-class classification task and considered out-of-domain because none of the pretraining datasets had tuberculosis labels.

\paragraph{CNS neoplasia diagnosis in MRI (CNS-MRI):}

The dataset of the CNS-MRI multi-class classification task \cite{BrainTumorMRIDataset2020} consisted of 7,023 2D slices derived from MRI scans categorized into four classes: glioma, meningioma, no tumor, and pituitary tumor. The slices originate from T1, T2 and FLAIR sequences and were selected by the authors of the dataset.
Prior to any evaluation, we randomly partitioned the images into a training set containing 70\% of the data and a test set containing the remaining 30\%.

\paragraph{Breast cancer classification in WSI (BC-Bach-WSI):}

The dataset of the BC-Bach-WSI multi-class classification task \cite{ARESTA2019122} was used for breast cancer classification in HE-stained whole histological slide images (WSIs). It consisted of four classes: normal, benign tumors, as well as in-situ and invasive carcinomas. The dataset was derived from 30 WSIs and was divided into image patches by the authors of the dataset. Each resulting image was annotated by two expert pathologists. From these images we used 76 normal tissues, 79 benign tumors, 80 in situ carcinomas, and 85 invasive cancers for training. For testing, we used 24 normal, 21 benign, 20 in situ, and 15 invasive images.

\paragraph{Breast cancer classification in microscopic images (BC-BHis-MIC):}

The dataset of the BC-BHis-MIC multi-class classification task focused on the binary classification of microscopic images from HE-stained breast tumors into benign lesions as opposed to malignant tumors \cite{7312934}. Benign lesions included adenosis, fibroadenoma, phylloides tumor, and tubular adenoma. The malignant tumor class contained 4 subtypes of invasive carcinoma: ductal carcinoma (currently referred to as "unspecific type"), lobular carcinoma, mucinous carcinoma, and papillary carcinoma. The authors of the dataset had achieved a maximum AUC at 200X magnification, which we also adopted for our analysis. The dataset contained 7,909 image patches from 82 patients, with 2,480 benign and 5,429 malignant images. To prepare for evaluation, we randomly divided the images into a training set (70\% of the data) and a test set (30\% of the data) prior to any analysis.

\paragraph{Organ segmentation in MRI (OrganSeg-MRI):}

The organ segmentation in MRI dataset OrganSeg-MRI was part of the AMOS22 challenge data \cite{ji2022amos} and consisted of 60 MRI volumes, split into 40 for training and 20 for testing. Each volume came with annotations for 15 different organs, spleen, right kidney, left kidney, gallbladder, esophagus, liver, stomach, aorta, inferior vena cava, pancreas, right adrenal gland, left adrenal gland, duodenum, bladder, and either the prostate or uterus.

\subsubsection*{Comparison of benchmark results}

We compared our results to the best study results previously reported for the seven target tasks. In this context, we determined the percentage of data that UMedPT required to achieve performance comparable to the external reference result. For each target task, the evaluation criteria from the respective reference papers were used. In 4 of 7 cases, the data set had to be split manually because the creators had not defined the test data. In these cases, manual splitting was performed only once to avoid bias.

\subsubsection*{Application of \networkName{} in an external multi-center validation of colorectal tumor diagnosis models}

\ifpreprint
\embargonotice{}
\fi

\section*{Data availability}

The raw data required to make the plots are distributed within the Source Data section. The training and evaluation data were obtained from their original repositories as listed below:

\begin{itemize}
    \item Amos22 \cite{ji2022amos} (organ segmentation in CT \& MRI): \url{https://amos22.grand-challenge.org/}
    \item Conic-WSI \cite{graham2021lizard} (cell detection): \url{https://conic-challenge.grand-challenge.org/}
    \item PICAL-MRI \cite{PICAI_BIAS} (prostate cancer classification) \url{https://pi-cai.grand-challenge.org/}: 
    \item Panda-WSI \cite{bulten2022artificial} (prostate tissue semantic segmentation \& classification): \url{https://www.kaggle.com/c/prostate-cancer-grade-assessment}
    \item VinBigData-CXR \cite{vinbigdata-chest-xray-abnormalities-detection} (Thorax pathology pathology detection): \url{https://www.kaggle.com/competitions/vinbigdata-chest-xray-abnormalities-detection}
    \item Crag-WSI \cite{DBLP:journals/corr/abs-1806-01963} (Colorectal tissue semantic segmentation): \url{https://github.com/XiaoyuZHK/CRAG-Dataset_Aug_ToCOCO}
    \item Brats2020-MRI \cite{menze2014multimodal} (brain semantic segmentation): \url{https://www.kaggle.com/datasets/awsaf49/brats20-dataset-training-validation}
    \item Avaniti-WSI \cite{arvaniti2018automated} (prostate multi-label classification): \url{https://doi.org/10.7910/DVN/OCYCMP}
    \item Cyto-WSI \cite{matek2021highly} (bone marrow single cell multi-class classification): \url{https://wiki.cancerimagingarchive.net/pages/viewpage.action?pageId=101941770}
    \item Chexpert-CXR \cite{irvin2019chexpert} (Thorax pathology multi-label classification): \url{https://stanfordaimi.azurewebsites.net/datasets/8cbd9ed4-2eb9-4565-affc-111cf4f7ebe2} \& \url{https://github.com/rajpurkarlab/cheXpert-test-set-labels}
    \item SIIM-CXR \cite{langershih2020} (pneumothorax semantic segmentation): \url{https://www.kaggle.com/competitions/siim-acr-pneumothorax-segmentation/data}
    \item ImageNet \cite{deng2009imagenet} (real world image classification): \url{https://www.image-net.org/download.php}
    \item RadImageNet \cite{mei2022radimagenet} (radiology multi-class classification): request access at \url{https://www.radimagenet.com/copy-of-home-1}
    \item COCO \cite{lin2015microsoft} (real world semantic segmentation \& object detection): \url{https://cocodataset.org/#download}
    \item CRC-WSI \cite{kather2019predicting} (colorectal cancer tissue classification): \url{https://zenodo.org/record/1214456}
    \item Pneumo-CXR \cite{KERMANY20181122} (pneumonia in pediatric cohort): \url{https://data.mendeley.com/datasets/rscbjbr9sj/3}
    \item Tuber-CXR \cite{jaeger2014two} (tuberculosis diagnosis in CXR): \url{https://www.kaggle.com/datasets/raddar/tuberculosis-chest-xrays-shenzhen}
    \item CNS-MRI \cite{BrainTumorMRIDataset2020} (CNS neoplasia diagnosis in MRI): \url{https://www.kaggle.com/datasets/masoudnickparvar/brain-tumor-mri-dataset}
    \item BC-Bach-WSI \cite{ARESTA2019122} (breast cancer classification in WSI): \url{https://iciar2018-challenge.grand-challenge.org/}
    \item BC-BHis-MIC \cite{7312934} (breast cancer classification in microscopic images): \url{https://web.inf.ufpr.br/vri/databases/breast-cancer-histopathological-database-breakhis/}
\end{itemize}

\section*{Code availability}

The code for reproducing our results is available at \url{https://zenodo.org/will/be/added/in/final/version}.
This archive contains the code for all experiments, including the corresponding archived version of the training framework.

If you intend to use \networkName{} or its training framework, we recommend that you use the latest version from the respective public projects hosted on Github.
The general training strategy is maintained as a Python package at: \url{https://github.com/FraunhoferMEVIS/will/be/added/in/final/version}.
The latest version of \networkName{} can be found at: \url{https://github.com/FraunhoferMEVIS/will/be/added/in/final/version}.

\bibliography{sn-bibliography}%

\section*{Acknowledgements}

This research was partly funded by the German ministry of education and research (BMBF) through the project SynDICAD (01IS21067C).

\section*{Author information}

RS and TN built pretraining and benchmark databases and implemented software.
RS conducted the experiments, analyzed the results, and wrote the manuscript, with feedback from all authors.
TN generated the plots and diagrams, and analyzed the results.
AL, DM, HH, FF, VS revised the manuscript.
FF: pathology advisor.
JL: coordinated the study.
FK: coordinated the study, radiology advisor.

\appendix

\renewcommand{\figurename}{Extended Data Fig.}
\renewcommand{\tablename}{Extended Data Table}
\renewcommand{\thesection}{S\arabic{section}}
\renewcommand{\thesubsection}{S\arabic{subsection}}

\setcounter{figure}{0}
\setcounter{table}{0}
\setcounter{section}{0}
\setcounter{subsection}{0}

\section*{Supplement}

\begin{table}
\footnotesize
\caption{\textbf{In-domain benchmark results.} The left pair of columns shows results with a frozen encoder, while the right pair shows results with fine-tuning (F1-scores reported in percentage). P-values, calculated independently for each dataset size, were always compared against the pretrained ImageNet model. \networkNameFixed{} was an ablation study where patch sizes remained constant, and \networkNameAffine{} was another ablation where layernorms had trainable parameters.}
\label{tab:resultsweakcombined}
\begin{tabular}{llllll}
\toprule
 &  & \multicolumn{2}{r}{frozen} & \multicolumn{2}{r}{finetune} \\
 &  & PPneumo-CXR & CRC-WSI & PPneumo-CXR & CRC-WSI \\
Size & Model &  &  &  &  \\
\midrule
\multirow[c]{4}{*}{1\%} & ImageNet & \makecell{61.68±12.55\%\\} & \makecell{80.81±0.54\%\\} & \makecell{85.54±2.32\%\\} & \makecell{91.84±1.70\%\\} \\
\cline{2-6}
 & \makecell{\networkName\\fixed} & \makecell{89.06±1.39\%\\p=4.54e-03} & \makecell{97.03±0.07\%\\p=3.13e-07} & \makecell{91.52±1.97\%\\p=2.19e-02} & \makecell{96.53±0.61\%\\p=2.58e-03} \\
\cline{2-6}
 & \makecell{\networkName\\affine} & \makecell{88.51±1.71\%\\p=5.54e-03} & \makecell{97.17±0.08\%\\p=2.96e-07} & \makecell{91.71±1.97\%\\p=1.74e-02} & \makecell{96.98±0.76\%\\p=7.02e-04} \\
\cline{2-6}
 & \networkName & \makecell{91.97±1.57\%\\p=3.09e-03} & \makecell{95.37±0.13\%\\p=8.26e-07} & \makecell{92.81±0.82\%\\p=2.18e-03} & \makecell{95.97±0.64\%\\p=2.32e-03} \\
\cline{1-6} \cline{2-6}
\multirow[c]{4}{*}{5\%} & ImageNet & \makecell{78.07±1.27\%\\} & \makecell{84.91±0.35\%\\} & \makecell{84.80±1.99\%\\} & \makecell{94.64±0.82\%\\} \\
\cline{2-6}
 & \makecell{\networkName\\fixed} & \makecell{89.49±1.29\%\\p=8.35e-05} & \makecell{96.96±0.11\%\\p=2.27e-07} & \makecell{91.34±1.84\%\\p=1.79e-03} & \makecell{96.57±0.34\%\\p=3.98e-03} \\
\cline{2-6}
 & \makecell{\networkName\\affine} & \makecell{88.24±1.44\%\\p=9.74e-05} & \makecell{96.96±0.15\%\\p=1.82e-08} & \makecell{91.04±2.01\%\\p=5.33e-04} & \makecell{96.78±0.30\%\\p=7.52e-03} \\
\cline{2-6}
 & \networkName & \makecell{93.46±0.65\%\\p=1.14e-06} & \makecell{95.45±0.14\%\\p=6.89e-07} & \makecell{92.02±2.03\%\\p=6.77e-04} & \makecell{95.68±0.70\%\\p=7.90e-02} \\
\cline{1-6} \cline{2-6}
\multirow[c]{4}{*}{10\%} & ImageNet & \makecell{79.05±1.54\%\\} & \makecell{86.13±0.35\%\\} & \makecell{85.06±6.84\%\\} & \makecell{95.44±0.56\%\\} \\
\cline{2-6}
 & \makecell{\networkName\\fixed} & \makecell{87.63±2.51\%\\p=1.02e-04} & \makecell{97.04±0.17\%\\p=2.65e-07} & \makecell{90.05±3.06\%\\p=8.43e-02} & \makecell{96.45±0.32\%\\p=5.16e-03} \\
\cline{2-6}
 & \makecell{\networkName\\affine} & \makecell{87.50±2.86\%\\p=2.06e-04} & \makecell{96.91±0.22\%\\p=1.99e-07} & \makecell{89.79±2.62\%\\p=8.19e-02} & \makecell{95.83±0.46\%\\p=1.25e-01} \\
\cline{2-6}
 & \networkName & \makecell{91.95±1.52\%\\p=2.16e-05} & \makecell{95.20±0.40\%\\p=7.47e-06} & \makecell{91.08±2.75\%\\p=3.12e-02} & \makecell{95.55±0.61\%\\p=3.93e-01} \\
\cline{1-6} \cline{2-6}
\multirow[c]{4}{*}{50\%} & ImageNet & \makecell{81.53±0.57\%\\} & \makecell{87.61±0.12\%\\} & \makecell{88.81±2.85\%\\} & \makecell{95.69±0.68\%\\} \\
\cline{2-6}
 & \makecell{\networkName\\fixed} & \makecell{89.56±0.76\%\\p=3.93e-05} & \makecell{96.83±0.13\%\\p=1.72e-08} & \makecell{90.50±3.02\%\\p=1.91e-02} & \makecell{95.70±0.68\%\\p=4.90e-01} \\
\cline{2-6}
 & \makecell{\networkName\\affine} & \makecell{89.65±0.64\%\\p=2.60e-05} & \makecell{96.70±0.12\%\\p=4.67e-11} & \makecell{92.16±1.04\%\\p=3.42e-02} & \makecell{95.78±0.76\%\\p=4.37e-01} \\
\cline{2-6}
 & \networkName & \makecell{90.52±0.35\%\\p=2.26e-07} & \makecell{95.59±0.13\%\\p=4.43e-08} & \makecell{90.58±1.94\%\\p=2.22e-01} & \makecell{95.01±0.69\%\\p=8.68e-01} \\
\cline{1-6} \cline{2-6}
\multirow[c]{4}{*}{100\%} & ImageNet & \makecell{82.18±0.94\%\\} & \makecell{87.68±0.07\%\\} & \makecell{90.34±1.77\%\\} & \makecell{95.16±0.59\%\\} \\
\cline{2-6}
 & \makecell{\networkName\\fixed} & \makecell{90.14±0.66\%\\p=8.95e-06} & \makecell{96.82±0.08\%\\p=4.69e-10} & \makecell{90.17±0.44\%\\p=5.83e-01} & \makecell{94.93±0.61\%\\p=6.80e-01} \\
\cline{2-6}
 & \makecell{\networkName\\affine} & \makecell{90.01±0.48\%\\p=2.18e-05} & \makecell{96.59±0.02\%\\p=4.60e-10} & \makecell{90.61±2.28\%\\p=4.42e-01} & \makecell{95.60±0.33\%\\p=9.77e-02} \\
\cline{2-6}
 & \networkName & \makecell{89.36±0.31\%\\p=3.40e-05} & \makecell{95.57±0.06\%\\p=1.43e-09} & \makecell{90.73±1.60\%\\p=3.65e-01} & \makecell{94.98±1.09\%\\p=5.94e-01} \\
\cline{1-6} \cline{2-6}
\bottomrule
\end{tabular}
\end{table}

\begin{table}
\footnotesize
\caption{\textbf{Out-of-domain benchmark with frozen encoder.} OrganSeg-MRI was evaluated using the mean Intersection over Union (mIoU) metric; all other tasks were analysed using F1-score (reported in percentage). P-values, calculated independently for each dataset size, were always compared against the pretrained ImageNet model. \networkNameFixed{} was an ablation study where patch sizes remained constant, and \networkNameAffine{} was another ablation where layernorms had trainable parameters.}
\label{tab:resultsstrongfrozen}
\begin{tabular}{lllllll}
\toprule
 &  & Tuber-CXR & OrganSeg-MRI & CNS-MRI & BC-BHis-MIC & BC-BACH-WSI \\
Size & Model &  &  &  &  &  \\
\midrule
\multirow[c]{4}{*}{1\%} & ImageNet & \makecell{37.88±5.86\%\\} & \makecell{67.78±0.00\%\\} & \makecell{31.49±9.61\%\\} & \makecell{47.94±7.05\%\\} & \makecell{32.98±3.58\%\\} \\
\cline{2-7}
 & \makecell{\networkName\\fixed} & \makecell{55.70±14.94\%\\p=2.27e-02} & \makecell{63.33±4.37\%\\p=9.44e-01} & \makecell{31.69±9.12\%\\p=4.73e-01} & \makecell{47.79±7.68\%\\p=5.20e-01} & \makecell{26.99±5.51\%\\p=8.87e-01} \\
\cline{2-7}
 & \makecell{\networkName\\affine} & \makecell{55.66±13.27\%\\p=1.75e-02} & \makecell{64.52±2.29\%\\p=9.77e-01} & \makecell{37.18±7.00\%\\p=7.53e-02} & \makecell{55.14±11.11\%\\p=6.43e-02} & \makecell{33.19±8.92\%\\p=4.82e-01} \\
\cline{2-7}
 & \networkName & \makecell{51.63±16.87\%\\p=4.38e-02} & \makecell{62.36±4.22\%\\p=9.69e-01} & \makecell{69.18±5.68\%\\p=7.19e-05} & \makecell{46.94±5.56\%\\p=6.24e-01} & \makecell{34.64±10.00\%\\p=3.94e-01} \\
\cline{1-7} \cline{2-7}
\multirow[c]{4}{*}{5\%} & ImageNet & \makecell{46.87±14.45\%\\} & \makecell{67.78±0.00\%\\} & \makecell{75.14±2.31\%\\} & \makecell{45.41±3.91\%\\} & \makecell{20.19±10.31\%\\} \\
\cline{2-7}
 & \makecell{\networkName\\fixed} & \makecell{67.62±3.28\%\\p=1.11e-02} & \makecell{66.93±0.72\%\\p=9.62e-01} & \makecell{80.21±2.85\%\\p=1.06e-03} & \makecell{55.47±4.68\%\\p=4.58e-03} & \makecell{26.26±9.31\%\\p=1.64e-01} \\
\cline{2-7}
 & \makecell{\networkName\\affine} & \makecell{68.30±6.40\%\\p=2.77e-02} & \makecell{67.37±0.15\%\\p=9.97e-01} & \makecell{78.15±3.69\%\\p=7.35e-03} & \makecell{61.37±9.69\%\\p=4.71e-03} & \makecell{33.10±13.31\%\\p=6.64e-02} \\
\cline{2-7}
 & \networkName & \makecell{76.67±16.65\%\\p=5.88e-03} & \makecell{68.30±1.13\%\\p=2.06e-01} & \makecell{86.80±0.76\%\\p=1.35e-04} & \makecell{74.23±2.77\%\\p=6.52e-06} & \makecell{29.78±12.12\%\\p=8.18e-02} \\
\cline{1-7} \cline{2-7}
\multirow[c]{4}{*}{10\%} & ImageNet & \makecell{50.42±11.04\%\\} & \makecell{67.78±0.00\%\\} & \makecell{80.03±0.84\%\\} & \makecell{57.23±13.31\%\\} & \makecell{39.90±8.90\%\\} \\
\cline{2-7}
 & \makecell{\networkName\\fixed} & \makecell{76.18±2.57\%\\p=1.97e-03} & \makecell{67.23±0.61\%\\p=9.27e-01} & \makecell{84.82±0.74\%\\p=8.57e-06} & \makecell{61.34±11.47\%\\p=2.13e-01} & \makecell{47.70±15.34\%\\p=7.10e-02} \\
\cline{2-7}
 & \makecell{\networkName\\affine} & \makecell{76.25±4.73\%\\p=1.87e-03} & \makecell{67.57±0.24\%\\p=9.22e-01} & \makecell{84.22±1.04\%\\p=7.93e-05} & \makecell{68.03±8.56\%\\p=1.86e-02} & \makecell{51.29±14.69\%\\p=5.00e-02} \\
\cline{2-7}
 & \networkName & \makecell{86.42±2.12\%\\p=9.65e-04} & \makecell{68.97±1.09\%\\p=4.73e-02} & \makecell{88.95±0.33\%\\p=5.82e-06} & \makecell{79.17±2.12\%\\p=9.91e-03} & \makecell{56.83±14.76\%\\p=4.62e-03} \\
\cline{1-7} \cline{2-7}
\multirow[c]{4}{*}{50\%} & ImageNet & \makecell{70.42±4.14\%\\} & \makecell{65.88±3.81\%\\} & \makecell{87.83±0.81\%\\} & \makecell{77.76±0.93\%\\} & \makecell{61.24±2.71\%\\} \\
\cline{2-7}
 & \makecell{\networkName\\fixed} & \makecell{85.03±1.10\%\\p=1.56e-03} & \makecell{68.11±0.13\%\\p=1.58e-01} & \makecell{91.65±0.28\%\\p=5.36e-04} & \makecell{84.12±0.26\%\\p=2.45e-05} & \makecell{70.63±6.97\%\\p=2.59e-02} \\
\cline{2-7}
 & \makecell{\networkName\\affine} & \makecell{84.18±1.32\%\\p=1.78e-03} & \makecell{67.84±0.01\%\\p=1.80e-01} & \makecell{90.68±0.38\%\\p=3.27e-03} & \makecell{85.18±1.42\%\\p=1.37e-03} & \makecell{75.92±3.07\%\\p=2.17e-04} \\
\cline{2-7}
 & \networkName & \makecell{90.50±0.91\%\\p=3.88e-04} & \makecell{72.48±0.08\%\\p=1.34e-02} & \makecell{92.95±0.33\%\\p=1.53e-04} & \makecell{86.94±1.14\%\\p=4.44e-05} & \makecell{77.97±0.96\%\\p=1.84e-04} \\
\cline{1-7} \cline{2-7}
\multirow[c]{4}{*}{100\%} & ImageNet & \makecell{67.99±1.11\%\\} & \makecell{67.78±0.00\%\\} & \makecell{89.05±0.11\%\\} & \makecell{82.30±0.58\%\\} & \makecell{72.90±1.49\%\\} \\
\cline{2-7}
 & \makecell{\networkName\\fixed} & \makecell{86.89±0.93\%\\p=6.22e-07} & \makecell{68.49±0.06\%\\p=9.02e-06} & \makecell{93.33±0.11\%\\p=3.37e-07} & \makecell{87.15±0.31\%\\p=1.32e-04} & \makecell{80.98±1.28\%\\p=1.09e-03} \\
\cline{2-7}
 & \makecell{\networkName\\affine} & \makecell{86.01±0.25\%\\p=1.76e-06} & \makecell{67.87±0.02\%\\p=3.23e-04} & \makecell{92.82±0.20\%\\p=1.19e-06} & \makecell{87.12±0.22\%\\p=4.94e-05} & \makecell{85.84±1.24\%\\p=1.87e-04} \\
\cline{2-7}
 & \networkName & \makecell{93.50±0.20\%\\p=1.12e-06} & \makecell{72.71±0.03\%\\p=2.21e-10} & \makecell{94.14±0.17\%\\p=1.60e-06} & \makecell{89.87±0.22\%\\p=1.11e-05} & \makecell{81.17±0.43\%\\p=2.39e-04} \\
\cline{1-7} \cline{2-7}
\bottomrule
\end{tabular}
\end{table}

\begin{table}
\footnotesize
\caption{\textbf{Out-of-domain benchmark with fine-tuned encoder.} OrganSeg-MRI was evaluated using the mean Intersection over Union (mIoU) metric; all other tasks were analysed using F1-score (reported in percentage). P-values, calculated independently for each dataset size, were always compared against the baseline. ImageNet served as the baseline, with \networkName{} representing our model. \networkNameFixed{} was an ablation study where patch sizes remained constant, and \networkNameAffine{} was another ablation where layernorms had trainable parameters.}
\label{tab:resultsstrongfinetune}
\begin{tabular}{lllllll}
\toprule
 &  & Tuber-CXR & OrganSeg-MRI & CNS-MRI & BC-BHis-MIC & BC-BACH-WSI \\
Size & Model &  &  &  &  &  \\
\midrule
\multirow[c]{4}{*}{1\%} & ImageNet & \makecell{58.21±9.50\%\\} & \makecell{66.74±3.90\%\\} & \makecell{69.24±2.81\%\\} & \makecell{68.96±6.90\%\\} & \makecell{36.44±5.12\%\\} \\
\cline{2-7}
 & \makecell{\networkName\\fixed} & \makecell{62.70±8.92\%\\p=2.98e-01} & \makecell{63.57±1.01\%\\p=9.32e-01} & \makecell{79.46±1.60\%\\p=3.81e-03} & \makecell{70.87±7.19\%\\p=1.29e-01} & \makecell{38.07±7.64\%\\p=3.29e-01} \\
\cline{2-7}
 & \makecell{\networkName\\affine} & \makecell{61.47±11.45\%\\p=3.53e-01} & \makecell{63.31±3.58\%\\p=9.97e-01} & \makecell{75.81±6.93\%\\p=2.08e-02} & \makecell{69.72±6.94\%\\p=3.71e-01} & \makecell{34.81±5.74\%\\p=6.75e-01} \\
\cline{2-7}
 & \networkName & \makecell{69.54±12.54\%\\p=1.11e-01} & \makecell{66.87±5.17\%\\p=4.79e-01} & \makecell{84.77±2.83\%\\p=4.98e-04} & \makecell{68.79±8.92\%\\p=5.29e-01} & \makecell{41.19±6.57\%\\p=1.34e-01} \\
\cline{1-7} \cline{2-7}
\multirow[c]{4}{*}{5\%} & ImageNet & \makecell{77.05±3.36\%\\} & \makecell{64.05±2.82\%\\} & \makecell{91.21±0.78\%\\} & \makecell{86.77±2.49\%\\} & \makecell{47.62±7.25\%\\} \\
\cline{2-7}
 & \makecell{\networkName\\fixed} & \makecell{83.44±1.85\%\\p=9.78e-03} & \makecell{69.43±3.33\%\\p=4.15e-02} & \makecell{93.57±0.36\%\\p=8.46e-04} & \makecell{88.45±1.63\%\\p=1.22e-02} & \makecell{43.02±6.84\%\\p=9.43e-01} \\
\cline{2-7}
 & \makecell{\networkName\\affine} & \makecell{82.44±2.88\%\\p=2.55e-02} & \makecell{69.63±3.80\%\\p=3.74e-02} & \makecell{93.01±0.80\%\\p=7.60e-04} & \makecell{86.39±1.30\%\\p=6.79e-01} & \makecell{42.42±8.40\%\\p=9.44e-01} \\
\cline{2-7}
 & \networkName & \makecell{92.51±2.36\%\\p=2.80e-04} & \makecell{74.70±3.92\%\\p=2.35e-03} & \makecell{93.50±0.74\%\\p=6.54e-03} & \makecell{87.41±1.96\%\\p=1.27e-01} & \makecell{47.61±12.52\%\\p=5.01e-01} \\
\cline{1-7} \cline{2-7}
\multirow[c]{4}{*}{10\%} & ImageNet & \makecell{81.17±2.49\%\\} & \makecell{67.49±2.72\%\\} & \makecell{93.53±1.43\%\\} & \makecell{89.29±1.55\%\\} & \makecell{59.84±8.95\%\\} \\
\cline{2-7}
 & \makecell{\networkName\\fixed} & \makecell{89.64±0.98\%\\p=1.38e-03} & \makecell{76.42±2.68\%\\p=7.69e-04} & \makecell{95.27±0.66\%\\p=1.12e-02} & \makecell{91.86±0.53\%\\p=1.07e-02} & \makecell{66.42±8.97\%\\p=6.17e-03} \\
\cline{2-7}
 & \makecell{\networkName\\affine} & \makecell{88.34±2.59\%\\p=9.94e-03} & \makecell{76.28±2.26\%\\p=1.17e-03} & \makecell{95.24±0.65\%\\p=7.05e-03} & \makecell{91.15±0.57\%\\p=6.92e-02} & \makecell{64.51±9.56\%\\p=6.65e-03} \\
\cline{2-7}
 & \networkName & \makecell{96.25±1.75\%\\p=5.86e-04} & \makecell{81.25±1.86\%\\p=5.75e-06} & \makecell{95.62±0.47\%\\p=7.18e-03} & \makecell{91.60±0.72\%\\p=3.11e-02} & \makecell{66.54±11.62\%\\p=7.55e-03} \\
\cline{1-7} \cline{2-7}
\multirow[c]{4}{*}{50\%} & ImageNet & \makecell{88.80±2.20\%\\} & \makecell{82.16±0.20\%\\} & \makecell{98.33±0.19\%\\} & \makecell{95.97±1.44\%\\} & \makecell{83.55±5.42\%\\} \\
\cline{2-7}
 & \makecell{\networkName\\fixed} & \makecell{93.01±0.66\%\\p=3.84e-03} & \makecell{86.51±0.16\%\\p=1.75e-06} & \makecell{98.30±0.16\%\\p=5.78e-01} & \makecell{97.21±0.49\%\\p=3.57e-02} & \makecell{89.58±3.20\%\\p=2.23e-02} \\
\cline{2-7}
 & \makecell{\networkName\\affine} & \makecell{92.61±0.24\%\\p=1.20e-02} & \makecell{86.34±0.08\%\\p=2.05e-06} & \makecell{98.24±0.15\%\\p=6.93e-01} & \makecell{96.65±0.85\%\\p=5.07e-02} & \makecell{87.87±3.09\%\\p=9.95e-02} \\
\cline{2-7}
 & \networkName & \makecell{95.44±0.72\%\\p=7.40e-04} & \makecell{87.02±0.13\%\\p=1.58e-06} & \makecell{98.63±0.11\%\\p=2.66e-02} & \makecell{97.20±0.80\%\\p=1.43e-02} & \makecell{88.81±2.79\%\\p=5.68e-02} \\
\cline{1-7} \cline{2-7}
\multirow[c]{4}{*}{100\%} & ImageNet & \makecell{90.13±4.98\%\\} & \makecell{85.86±0.39\%\\} & \makecell{98.98±0.13\%\\} & \makecell{98.39±0.35\%\\} & \makecell{92.99±2.24\%\\} \\
\cline{2-7}
 & \makecell{\networkName\\fixed} & \makecell{94.13±0.82\%\\p=9.53e-02} & \makecell{87.27±0.25\%\\p=2.60e-03} & \makecell{99.27±0.06\%\\p=1.25e-02} & \makecell{98.83±0.19\%\\p=5.34e-02} & \makecell{95.54±1.33\%\\p=6.15e-02} \\
\cline{2-7}
 & \makecell{\networkName\\affine} & \makecell{93.12±0.52\%\\p=1.48e-01} & \makecell{87.37±0.36\%\\p=2.70e-04} & \makecell{99.15±0.20\%\\p=6.26e-02} & \makecell{98.23±0.45\%\\p=6.47e-01} & \makecell{94.54±1.61\%\\p=2.09e-01} \\
\cline{2-7}
 & \networkName & \makecell{93.92±0.46\%\\p=1.11e-01} & \makecell{88.14±0.14\%\\p=1.05e-04} & \makecell{99.27±0.15\%\\p=1.65e-02} & \makecell{98.38±0.60\%\\p=5.07e-01} & \makecell{92.67±2.10\%\\p=5.64e-01} \\
\cline{1-7} \cline{2-7}
\bottomrule
\end{tabular}
\end{table}

\begin{figure}
    \centering

    \begin{minipage}[b]{0.9\textwidth}
        \stackinset{l}{1pt}{t}{1pt}{\color{black}\bfseries a}{\includegraphics[width=\textwidth]{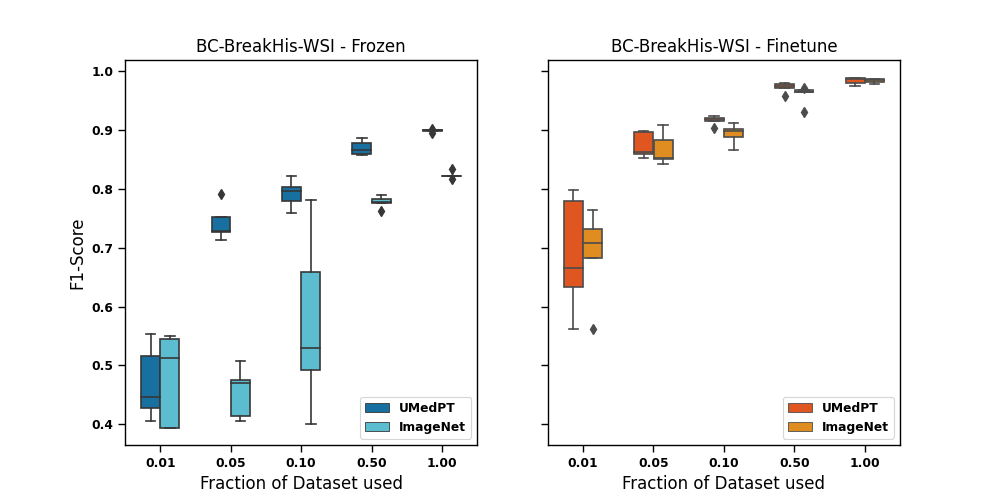}}
        \label{fig:alloodplots3}
    \end{minipage}

    \begin{minipage}[b]{0.9\textwidth}
        \stackinset{l}{1pt}{t}{1pt}{\color{black}\bfseries b}{\includegraphics[width=\textwidth]{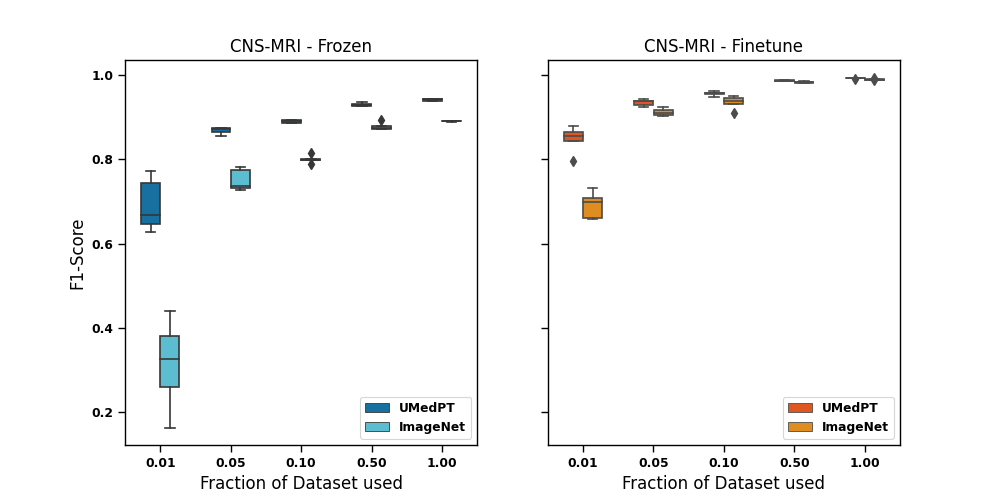}}
        \label{fig:alloodplots4}
    \end{minipage}

    \begin{minipage}[b]{0.9\textwidth}
        \stackinset{l}{1pt}{t}{1pt}{\color{black}\bfseries c}{\includegraphics[width=\textwidth]{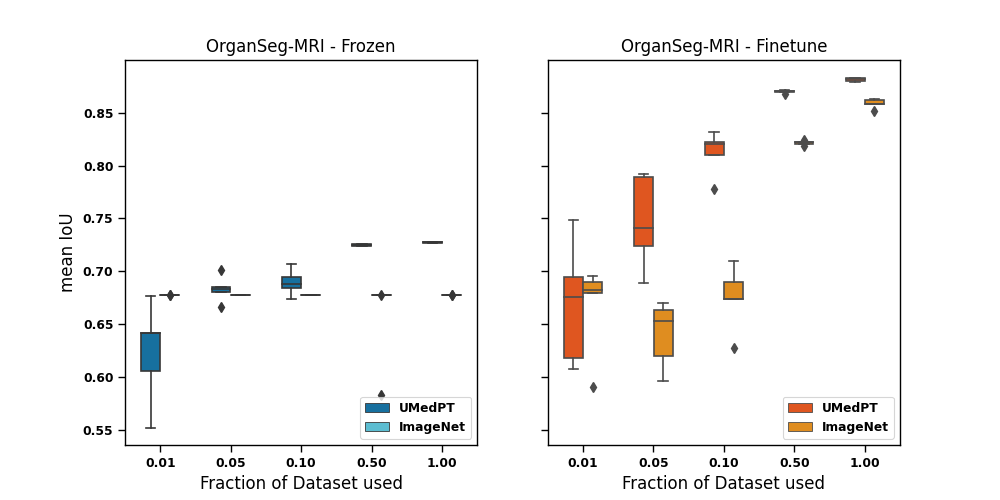}}
        \label{fig:alloodplots5}
    \end{minipage}
    \caption{\textbf{Results of remaining out-of-distribution tasks.} \textbf{a} BC-BHis-MIC \textbf{b} CNS-MRI. \textbf{c} OrganSeg-MRI.}
    \label{fig:alloodplots}
\end{figure}

\clearpage

\subsection{State of the art for the tasks}
\label{sec:sotadetails}

Each item in the list corresponds to one reference result in Table \ref{tab:datafraction}.

\begin{itemize}
    \item \textbf{CRC-WSI:} Over 94\% accuracy in the nine-class classification problem was achieved with ImageNet transfer learning \cite{kather2019predicting}.
    \item \textbf{Pneumo-CXR:} The dataset creators reported a diagnostic accuracy of 92.8\%, a sensitivity of 93.2\%, and a specificity of 90.1\% \cite{KERMANY20181122}.
    \item \textbf{Tuber-CXR:} Radiologists achieved an accuracy of 82\% \cite{jaeger2013automatic}. Specialized representation learning for CXR achieved an AUC of 98.0\%, with their corresponding ImageNet baseline reaching an AUC of 94.5\% \cite{zhou2022generalized}.
    \item \textbf{CNS-MRI:} Classifiers with different data splits from the Kaggle community achieved over 96\% accuracy.
    \item \textbf{BC-Bach-WSI:} An accuracy of 87\% was achieved using external validation on a challenge server \cite{ARESTA2019122}
    \item \textbf{BC-BHis-MIC:} The dataset creators reported an accuracy between 80\% and 85\% \cite{7312934}. The highest reported F1-score was 88\% \cite{chhipa2022magnification}.
\end{itemize}

\end{document}